\documentclass[11pt]{article}

\usepackage[final]{acl}
\usepackage{multirow}
\usepackage{times}
\usepackage{latexsym}
\usepackage{amsmath}
\usepackage[T1]{fontenc}

\usepackage[utf8]{inputenc}

\usepackage{microtype}

\usepackage{inconsolata}

\usepackage{graphicx}
\usepackage{float} 
\usepackage{booktabs}

\title{PragMatch: A Benchmark for Separating Pragmatic Incongruity from Cross-Modal Mismatch in Vision Language Models}

\author{
 \textbf{Zhanna Mukhametsharip\textsuperscript{1}},
 \textbf{Vera Demberg\textsuperscript{1,2}},
 \textbf{Varsha Suresh\textsuperscript{2}}
\\
\\
 \textsuperscript{1}Saarland University, Germany \\
\textsuperscript{2}Max Planck Institute for Informatics, Germany \\
Saarland Informatics Campus
\\
 \small{
   \url{zhmu00001@stud.uni-saarland.de},  \url{vera@lst.uni-saarland.de}, \url{vsuresh@mpi-inf.mpg.de}
 }
}

\begin{document}
\maketitle  
 
\begin{abstract} 

Large Vision--Language Models (LVLMs) have demonstrated strong performance on multimodal benchmarks, yet it remains unclear whether they genuinely reason about relationships between images and text or rely on superficial correlations, known as shortcut learning. This question is particularly important for multimodal sarcasm detection, where successful prediction depends on recognizing pragmatic incongruity rather than treating sarcasm as simple image--text mismatch. We introduce PragMatch, a controlled benchmark of 3,000 image–text pairs derived from MMSD2.0, including original sarcastic examples and constructed literal and hard-negative pairs. Beyond standard accuracy, we evaluate at the relation level, requiring joint correctness across an image's matched conditions, and report AUROC. We identify influential shortcut cues through systematic masking and evaluate their impact through targeted injection experiments. Our results show that LVLM predictions are sensitive to lexical, OCR-derived and stylistic cues, with injected surface signals causing substantial changes in model predictions despite unchanged underlying image--text relationships. Our findings reveal limitations in evaluated LVLMs while PragMatch provides a systematic testbed for evaluating multimodal pragmatic reasoning beyond surface-level image–text alignment. The benchmark and code are available at
\url{https://zmukhametsharip.github.io/pragmatch-project-page/}.

\end{abstract}

\section{Introduction}

Multimodal sarcasm detection is usually formulated as binary classification and evaluated with Accuracy and Macro F1 \citep{cai2019multi, qin2023mmsd2}. These metrics ask whether a model assigns the correct label to each image and caption pair on its own, and they allow direct comparison across models. They do not ask why the model considers a pair incongruent. In particular, they do not test whether a model separates incongruity that carries communicative intent from incongruity that is arbitrary. Dataset work has reduced one source of this problem by removing explicit sarcasm markers and correcting annotation errors \citep{qin2023mmsd2}, but the evaluation protocol itself still scores every example in isolation.

This matters because pragmatic incongruity is not the same as general inconsistency between an image and a caption. A sarcastic caption and a caption drawn from an unrelated image can both conflict with the visual content, yet only the first conflict is intended and interpretable. Recognising sarcasm therefore requires a model to identify which relation holds between the two modalities, not simply to notice that they disagree. A model that treats any inconsistency as sarcasm, or that rejects any inconsistency as noise, can still score well when each example is judged alone.

Recent diagnostic work suggests that this concern is well founded. NaturalBench shows that imbalanced answer distributions let models answer visual questions from language priors, and it responds by pairing each question with images that require opposite answers \citep{li2024naturalbench}. \citet{kamath2024hardpositivetruthvisionlanguage} show that models trained on hard negatives become sensitive to the presence of an edit rather than to its meaning, and that robustness requires sensitivity to meaning changing variation together with invariance to meaning preserving variation. PunchBench reports that humour and sarcasm questions are often answerable from the caption alone \citep{ouyang2025punchbench}, and \citet{chi2025chimera} show that VLMs exceed chance on diagram reasoning with the image removed. Work on multimodal sarcasm has begun to address pragmatic reasoning directly \citep{mustreason2025, ironic2025}, but no existing benchmark isolates the discrimination of communicative relations from the detection of mismatch in general.

We build such a benchmark from 3{,}000 image and caption pairs derived from MMSD2.0 \citep{qin2023mmsd2}. Each source image is paired with three matched captions that hold the visual content fixed and vary only the communicative relation, namely Pragmatic Incongruity (P), Non-pragmatic Mismatch (M) and Literal Congruity (L). The mismatched captions are hard negatives retrieved from visually similar images with CLIP \citep{radford2021learning} and filtered with a relational similarity model \citep{nguyen2025relational}, so that they are plausible at the surface level rather than trivially unrelated. Alongside the standard classification metrics we report paired and grouped accuracy in the spirit of NaturalBench \citep{li2024naturalbench}, which give credit only when a model classifies all conditions of a source image correctly, together with AUROC to separate representation quality from decision bias. 

The results reveal a gap between standard accuracy and relation-level understanding. Under zero-shot prompting, these results reveal an imbalance across conditions: some models perform well on pragmatic pairs but poorly on non-sarcastic mismatches, while others show the opposite pattern, resulting in below-chance paired accuracy. CoT prompting improves sarcastic-pair recognition across all models but also increases false sarcasm detection for some models. Score-level analysis shows that relational information can be present without reliable decisions (Qwen2.5-VL-7B: 90.0\% AUROC vs. 13.4\% paired accuracy). Shortcut perturbations further demonstrate sensitivity to surface cues,  with OCR-based modifications producing the largest performance changes, rather than robust image--caption relation modeling.
\paragraph{Contributions.} We introduce PragMatch\footnote{We release the PragMatch benchmark and code at project page:
\url{https://zmukhametsharip.github.io/pragmatch-project-page/}.}, a controlled benchmark of 3{,}000 image--text pairs that isolates pragmatic incongruity from non-pragmatic mismatch and literal congruity relations, keeping the image fixed. We propose relation-level evaluation via relation substitution, surface perturbation, and modality ablation, revealing that strong single-condition performance can coexist with paired or group accuracy below chance. We diagnose the mechanisms behind these failures through masking and injection interventions over lexical, style and OCR cues, which leave the underlying relation and the gold label unchanged and therefore require stable predictions.

\section{Related Work}

\subsection{Controlled Benchmarks for LVLMs}

Strong performance on standard vision-language benchmarks does not necessarily reflect genuine multimodal reasoning, since models may exploit language priors, dataset biases, or superficial correlations rather than grounding predictions in the relationship between modalities. This has motivated controlled benchmarks that test whether models capture that relationship. A major line of work evaluates compositionality by constructing hard negatives, modifying captions or visual inputs to introduce compositional changes \citep{yuksekgonul2023visionlanguagemodelsbehavelike, ma2023crepe, hhsieh2023sugarcrepefixinghackablebenchmark, doveh2023teachingstructuredvisionlanguageconcepts}. Strong performance on these benchmarks is not decisive, since models may succeed simply by detecting a modification, without determining whether it changes the underlying meaning.

\citet{kamath2024hardpositivetruthvisionlanguage} demonstrate this limitation by introducing hard positives, meaning-preserving edits such as synonym substitution or reordering, and show that models fine-tuned on hard negatives erroneously assign lower scores to these semantically equivalent inputs. Robustness therefore requires two properties: sensitivity to meaning-changing variation and invariance to meaning-preserving variation. NaturalBench \citep{li2024naturalbench} identifies a similar failure in visual question answering, where answer imbalance enables language-only models to partially solve benchmarks. It addresses this by pairing each question with two images requiring opposite answers. PunchBench \citep{ouyang2025punchbench} extends this paradigm to multimodal humor and sarcasm, showing that models often answer from the caption alone and that synonym or antonym substitution substantially alters predictions even when the image and caption relationship is unchanged.

Existing benchmarks vary semantic compatibility, asking whether an image and caption are compatible or whether an edit changes literal meaning. We instead control the type of image--text relation, asking whether a literally incompatible caption forms an intentional pragmatic relation or an arbitrary mismatch.

\subsection{Shortcut Learning and Diagnostic Interventions}

Beyond benchmark-level controls, recent work intervenes directly on model inputs to identify which cues drive predictions. Shortcut behavior has been observed across visual question answering \citep{si2022language}, vision-language representation learning \citep{bleeker2024demonstrating}, visual correspondence \citep{shahgir2026vlms} and diagram reasoning \citep{chi2025chimera}. The responsible cues originate in both modalities, including language priors \citep{si2022language}, lexical and stylistic patterns \citep{Du2023, buzeta2026seeing}, semantic associations \citep{shahgir2026vlms} and background visual biases \citep{xu2025overcoming}. \citet{chi2025chimera} demonstrate the problem directly by showing that VLMs perform above chance on diagram reasoning with the image removed entirely. This reflects the broader concern raised by \citet{kamath2024hardpositivetruthvisionlanguage}, that models may respond to the presence of a cue rather than to its meaning, and reliance on particular shortcuts depends on both the training data and the modality in which the cue appears \citep{buzeta2026seeing, xu2025overcoming}. These findings motivate diagnosis through modality ablation and cue perturbations.

\subsection{Multimodal Sarcasm and Pragmatic Incongruity}
Multimodal sarcasm is a natural testbed for evaluating pragmatic reasoning, as it combines cross-modal contradiction with communicative intent. Early social media datasets contained explicit cues such as hashtags and emojis correlated with sarcasm labels \citep{schifanella2016detecting, cai2019multi}; MMSD2.0 reduced these shortcuts by removing markers and correcting annotation errors \citep{qin2023mmsd2}. Shortcut reliance nonetheless persists: \citet{jia2024debiasing} identify spurious correlations between text features and sarcasm labels and propose contrastive debiasing, while other work explores incongruity-aware representation learning and debiasing objectives \citep{guo2025micl, wen2023dip}. These methods improve prediction but do not diagnose which cross-modal relations models actually use. Closer to our setting, MUStReason benchmarks pragmatic reasoning in video-language models \citep{mustreason2025}, and IRONIC separates referential, analogical, and pragmatic incongruity through a coherence-aware framework \citep{ironic2025}. We complement this direction with a focus on diagnosis and benchmark construction, using controlled masking and injection to isolate surface-level cues and test whether LVLMs distinguish pragmatic incongruity from non-pragmatic image--text mismatch.
\section{PragMatch Benchmark}


\begin{figure*}[t]
    \centering
    \includegraphics[width=0.85\textwidth]{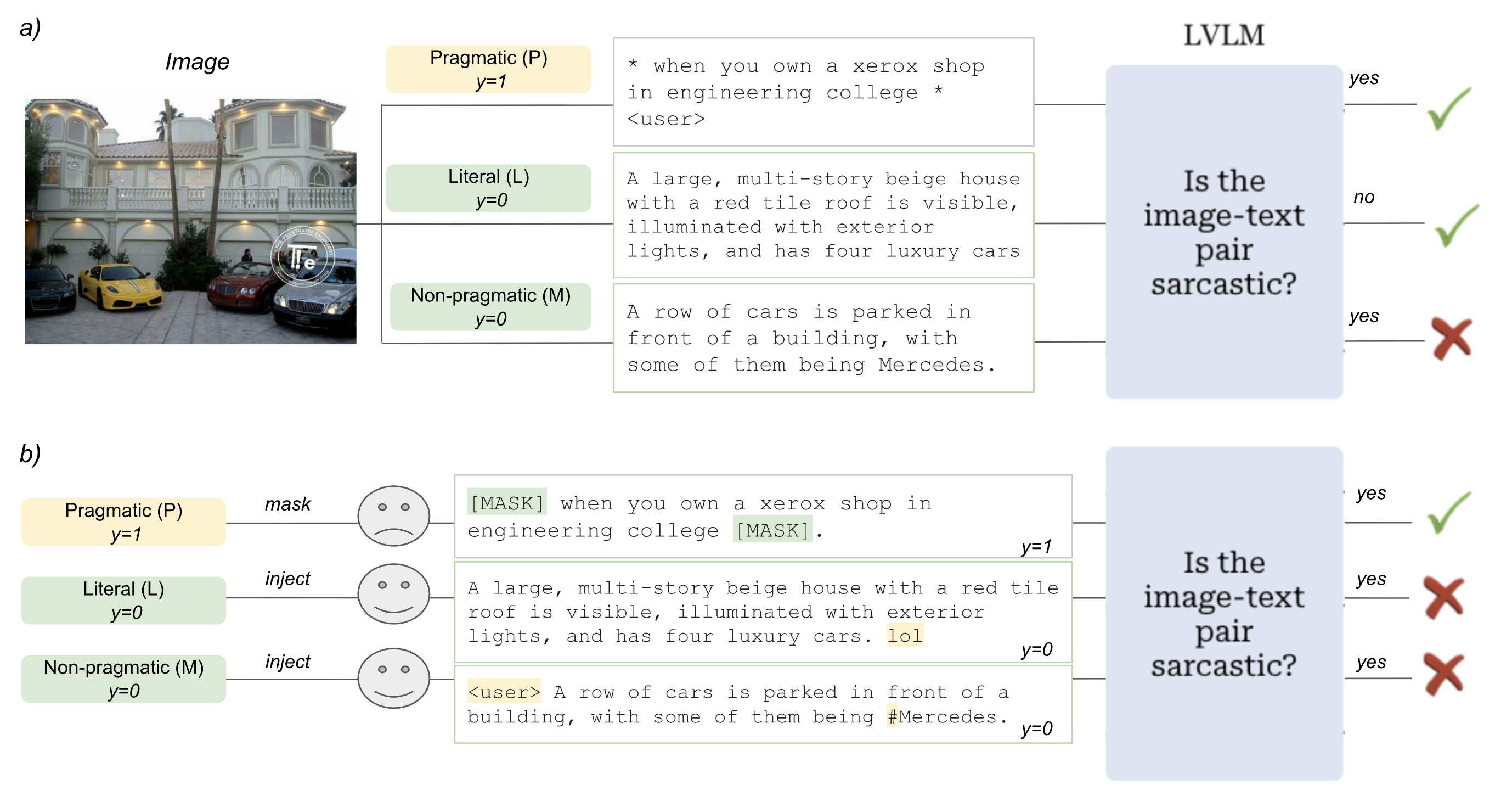}
    \caption{PragMatch benchmark and shortcut perturbation pipeline. (a) The benchmark compares three types of image–text relationships for the same image: sarcastic pairs with conflicting intended meaning (P), literal pairs that match the visual content (L), and mismatched pairs whose captions are drawn from visually similar but different images (M). (b) The surface perturbation framework tests whether LVLMs rely on superficial textual cues by masking and injecting controlled text perturbations.}
    \label{fig:frame}
\end{figure*}

\subsection{Overview}
Figure~\ref{fig:frame}a illustrates our evaluation protocol using a single source image paired with three captions. The  Pragmatic Incongruity (P) caption is sarcastic and requires understanding the intended meaning beyond the observable content. The Non-Pragmatic Mismatch (M) caption creates image-text mismatch without communicative intent, while the  Literal (L) caption describes the image content directly. This comparison shows that image-text inconsistency alone is insufficient for sarcasm detection; models must distinguish pragmatic incongruity from ordinary mismatch. 
Formally, we denote the source image as $x$, the associated caption set as $T(x)$, the gold label as $y \in \{0,1\}$, the model prediction as $\hat{y}$, and the continuous sarcasm score as $s$. Let $x$ denote a source image and $T(x)$ its associated caption set:
\[
T(x)=\{t_{\mathrm{P}}, t_{\mathrm{L}}, t_{\mathrm{M}}\}\]
Each image-caption pair $(x,t)$ is assigned a gold sarcasm label $y \in \{0,1\}$, where $y=1$ indicates sarcasm. The evaluation unit is a single image-caption pair. Following HallusionBench~\citep{guan2024hallusionbench}, the model receives an image and caption and produces a binary\textit{ Yes/No} verdict indicating whether the pair is sarcastic. The model outputs a binary prediction $\hat{y}$ and a continuous sarcasm score $s$ is computed from the log-probabilities of the verdict tokens (Eq.~\ref{eq:score}):
\begin{equation}
s = \log \sum_{\hat{y} \in \mathcal{Y}_{\text{yes}}} P(\hat{y})
- \log \sum_{\hat{y}\in \mathcal{Y}_{\text{no}}} P(\hat{y})
\label{eq:score}
\end{equation}
where $\mathcal{Y}_{\text{yes}}$ and $\mathcal{Y}_{\text{no}}$ denote the token sets corresponding to “Yes” and “No”. Positive $s$ indicates sarcasm and negative $s$ indicates non-sarcasm, enabling threshold-independent evaluation.


\subsection{Design Principle}
Paired accuracy P--M is the core contrast for testing pragmatic understanding: both conditions contain image–text mismatch, but only P  conveys sarcastic intent. To further analyze what evidence models rely on, we evaluate three controlled settings: 1) \textit{Relation substitution} changes the image-text relationship and therefore the gold label. For a fixed image, replacing a \textit{P} caption with an \textit{M} or \textit{L} changes the expected sarcasm decision; a model that captures the underlying relationship should update its prediction accordingly. 2) \textit{Surface perturbation} changes only the textual form while preserving the intended relation, keeping the label fixed. 3) \textit{Modality ablation} removes the image while keeping the caption unchanged; a model relying on genuine multimodal understanding should experience a performance drop.  These three evaluations follow the same idea as the hard positive analysis of \citet{kamath2024hardpositivetruthvisionlanguage}: models should respond to changes that alter the underlying meaning, while remaining stable under meaning-preserving perturbations.  Combining relation substitution and surface perturbation motivates the $P^{-}$--$M^{+}$ setting, where surface-level cues point toward the wrong decision, while the underlying pragmatic relation should stay the same.

\subsection{Benchmark Construction}
\label{sec:benchmark_construction}
We construct a 3,000 image-text pair evaluation set with 1,000 examples per condition: Pragmatic Incongruity ($P$), Literal ($L$), and Non-Pragmatic Mismatch ($M$).
\paragraph{Source data.}
The pragmatic condition ($t_{\text{P}}$) is retrieved from MMSD2.0 \citep{qin2023mmsd2}. We randomly sample image--text pairs with a sarcasm label of $y=1$ from the MMSD2.0 training set annotations to obtain the original sarcastic examples. 
We use MMSD2.0 because it improves over earlier sarcasm datasets by reducing reliance on explicit sarcasm markers and correcting annotation inconsistencies, making it more suitable for evaluating whether models capture pragmatic meaning rather than surface-level cues \citep{qin2023mmsd2}.


\paragraph{Literal captions.}
Literal captions $t_{\text{L}}$ serve as a congruent non-sarcastic control. They are generated with Gemma-3-27B-IT \citep{gemmateam2025gemma3technicalreport} using a constrained prompt that describes only observable image content (objects, actions, and visible text), while avoiding inferred context or intentions. Generation uses greedy decoding ($\texttt{do\_sample=False}$) with a 40-token limit to ensure consistent visual descriptions.  

\paragraph{Hard negatives.}
We construct $t_{\text{M}}$  captions through semantic filtering. CLIP \citep{radford2021learning} retrieves the five nearest visual neighbors for each image; duplicates of the original literal caption are removed. These candidates are then filtered using RelSim \citep{nguyen2025relational}, a metric designed to capture relational similarity beyond surface-level visual attributes.  Candidates with low-to-moderate relational similarity (0.2--0.4) measured by RelSim, or closest to this range, are retained, preserving visual plausibility while reducing relational correspondence. The resulting negatives are visually similar but semantically mismatched, creating challenging controls for multimodal reasoning. The caption is the literal description of another image generated with LLaVA-1.5-7B \cite{liu2023improvedllava}.

\paragraph{Surface perturbations.}
We study three families of surface cues that may introduce shortcuts into sarcasm prediction: \textit{lexical cues}, which modify explicit lexical markers (e.g.,"ironic"); \textit{style cues}, which modify social-media writing patterns (e.g., hashtags, user mentions); and \textit{OCR cues}, which modify textual evidence embedded in the image. Cue selection draws on these sources: <user> is the highest-weighted TF-IDF term distinguishing $P$ from $M$, appearing in 37.6\% of sarcastic captions; lexical and hashtag markers follow prior work \citep{schifanella2016detecting, cai2019multi, qin2023mmsd2} and their frequency in the source data. Candidates were manually screened for meaning preservation and validated by human annotation  (Section \ref{sec:quality_control}). For each cue family, we apply two complementary transformations: \textit{masking}, which removes existing cues, and \textit{injection}, which adds cues to examples with reliable gold labels. Masking removes potential shortcut cues from sarcastic examples ($P$), while injection adds misleading cues to non-sarcastic examples ($M$, $L$), with both interventions preserving the gold label. Table~\ref{tab:injection_temp} presents examples of the masked cues and the injection templates.
By design, these interventions modify the caption without altering the image-caption relationship or the gold label. Following the sensitivity--invariance principle of hard-positive evaluation \citep{kamath2024hardpositivetruthvisionlanguage}, a robust model should remain invariant to these surface edits rather than changing its prediction based only on shortcut cues. 
\subsection{Quality Control}
\label{sec:quality_control}
 

We perform two analyses to validate the benchmark. First, we test whether caption-only shortcuts can solve the task using a text-only BERT classifier. Using source-image-level splits, a frozen BERT classifier trained only on captions achieves 99.6\% item accuracy and 99.3\%  $P-M$ paired accuracy. This separation partly reflects caption provenance, as $P$ consists of pre-tokenised social-media text while $M$ and $L$ are model-generated. Strong text-only performance reveals caption-level biases that models may exploit as shortcuts, a pattern also observed in multimodal benchmarks such as PunchBench, where punchline questions are often answerable from text alone \citep{ouyang2025punchbench}. Our masking and injection interventions (Section~\ref{sec:benchmark_construction}) edit the same caption in place, holding condition, image, and gold label fixed, so stylistic differences between $P$, $M$, and $L$ cannot explain prediction changes under perturbation.

Second, we measure human performance and annotation consistency to assess task difficulty and label reliability. Two annotators evaluated 400 image--caption instances to verify condition labels. Human performance reached 85.8\% item accuracy and 71.5\% paired accuracy. The main sarcasm annotation achieved Cohen's $\kappa=0.518$ with 77.5\% agreement, while validity checks showed high agreement for literal support (91.0\%) and mismatch sarcasm plausibility (100\%). These results confirm that the benchmark conditions are interpretable, with moderate agreement on the main sarcasm judgement and high agreement on the validity checks. The perturbations were further validated: 150 injected items (50 per cue family) were judged by three annotators for whether the edit changes the gold label or the pragmatic interpretation. Majority judgements give 94.0\% gold-label preservation (lexical 96.0, style 90.0, OCR 96.0) and 98.7\% interpretation preservation.
\subsection{Overall Benchmark Statistics}
Our benchmark contains 3,000 image--caption pairs, with 1,000 examples per relation condition. The core $P-M$ comparison is constructed as a balanced subset, ensuring that accuracy and F1 scores are not affected by class imbalance. For shortcut analysis, injection is applied to all 1,000 target examples, while masking is applied only when the relevant cue is present. Valid masking subset contains $n=19$ lexical examples, $n=638$ style examples, and $n=506$ OCR masking examples. The lexical masking subset is small because explicit sarcasm markers occur rarely in MMSD2.0. Results on this subset therefore show that such cues can influence predictions, rather than how often they do.


\section{Experimental Setup}
\subsection{Evaluation Metrics}
We report several metrics to characterize sarcasm recognition and robustness. \textit{Condition accuracy} measures performance on each relation type ($P$, $M$ and $L$) separately. We report accuracy and binary \textit{F1}, with $P$ as the positive class, on the balanced $P$--$M$ subset to evaluate sarcasm discrimination across classes. Following grouped evaluation protocols in NaturalBench~\citep{li2024naturalbench}, we define stricter relation-level accuracies over captions derived from the same image: Paired accuracy  $P$--$M$ requires both $P$ and $M$ correct for the same image; grouped accuracy $P$--$M$--$L$ requires all three captions to be correctly classified. To evaluate shortcut robustness, we define $P$--$P^-$ as correct if both original and masked pragmatic captions are classified as sarcastic, and $M$--$M^+$ as correct if both original and mismatch-injected captions are classified as non-sarcastic. \textit{AUROC} is computed from continuous sarcasm scores to measure threshold-independent class separation. Because caption form partly separates the conditions (Section \ref{sec:quality_control}), we treat $P$--$M$ as an upper bound on relation sensitivity and use $P^-$--$M^+$, grouped accuracy, and AUROC as primary measures of pragmatic grounding.

\subsection{Models}
We evaluate four open-source vision--language models: 
 InternVL2.5-8B \citep{chen2024internvl25}, 
Qwen2.5-VL-7B-Instruct \citep{qwen2025qwen25vl}, 
LLaVA-v1.6-7B \citep{liu2024llavanext}, 
and Idefics2-8B \citep{laurencon2024idefics2}. 
These models represent diverse LVLM architectures and training approaches, allowing us to assess shortcut reliance across model families. All models are evaluated with deterministic decoding, on a Tesla T4 GPU and an NVIDIA GPU on HTCondor, using the same prompting configuration.
\subsection{Prompting}
 \paragraph{Zero-shot.} The prompt asks the model to determine whether the caption is sarcastic with respect to the image and return a binary \textit{Yes}/\textit{No} verdict.
 \paragraph{Chain-of-Thought.} The prompt employs a structured chain-of-thought adapted from \citet{Yue_Mao_Shi_Cambria_2026}. The reasoning chain consists of five steps: (1) describe the image and its meaning, (2) describe the caption and its meaning, (3) explain the joint meaning, (4) infer the author's intended message, and (5) predict whether the caption is sarcastic. Compared with \citet{Yue_Mao_Shi_Cambria_2026}, our adaptation replaces explicit sentiment reasoning with semantic meaning and intended-message inference to better reflect pragmatic sarcasm understanding (see Appendix ~\ref{appendix:prompts}). Outputs without a valid \textit{Yes}/\textit{No} prediction are treated as unparseable and excluded from evaluation.
\section{Results}

\begin{table*}[h]
\centering
\resizebox{\linewidth}{!}{
\begin{tabular}{ll|cc|ccc|cc|c}
\hline
& & \multicolumn{2}{c|}{\textbf{Standard Eval.}} & \multicolumn{3}{c|}{\textbf{Per-Condition Acc.}} & \multicolumn{2}{c|}{\textbf{Relation-Level Acc.}} & \\
\textbf{Model} & \textbf{Prompt} & \textbf{Acc} & \textbf{F1} & \textbf{P} & \textbf{M} & \textbf{L} & \textbf{P--M} & \textbf{P--M--L} &
\shortstack[c]{\textbf{Acc. Gap}\\{\scriptsize(Acc $-$ P--M--L)}} \\
\hline
\textit{Random Chance} & & \textit{50.0} & \textit{50.0} & \textit{50.0} & \textit{50.0} & \textit{50.0}
& \textit{25.0} & \textit{12.5} & \textit{37.5} \\
\hline
\multirow{2}{*}{InternVL2.5-8B} & Zero-Shot
& 52.9 & 14.0 & \phantom{0}7.7 & 98.0 & 94.4
& \phantom{0}7.4 & \phantom{0}6.0 & \textcolor{red}{46.9} \\
& CoT
& 53.3 & 58.9 & 65.4 & 39.9 & 83.0
& 25.8 & 23.3 & \textcolor{red}{30.0} \\
\hline
\multirow{2}{*}{Qwen2.5-VL-7B} & Zero-Shot
& 56.6 & 24.3 & 13.9 & 99.3 & 84.8
& 13.4 & \phantom{0}5.6 & \textcolor{red}{51.0} \\
& CoT
& 66.2 & 67.7 & 70.8 & 61.5 & 58.0
& 38.3 & 26.4 & \textcolor{red}{39.8} \\
\hline
\multirow{2}{*}{LLaVA-v1.6-7B} & Zero-Shot
& 74.2 & 70.8 & 62.5 & 85.9 & 43.1
& 50.6 & 19.0 & \textcolor{red}{55.2} \\
& CoT
& 63.3 & 72.5 & 96.7 & 29.9 & 37.6
& 28.1 & 17.4 & \textcolor{red}{45.9} \\
\hline
\multirow{2}{*}{Idefics2-8B} & Zero-Shot
& 53.5 & 18.1 & 10.4 & 96.7 & 39.2
& 10.2 & \phantom{0}2.7 & \textcolor{red}{50.8} \\
& CoT
& 57.8 & 57.4 & 56.9 & 58.7 & 47.1
& 28.8 & 15.8 & \textcolor{red}{42.0} \\
\hline
\textit{{Human Eval.}} & --
& 85.8 & 82.8 & 71.5 & 100.0 & 93.5
& 71.5 & \phantom{0}67.5 & 18.3 \\
\hline
\end{tabular}}
\caption{
Standard vs.\ relation-level evaluation on the $P$/$M$/$L$ contrast set.
$P$--$M$ and $P$--$M$--$L$ require joint correctness across matched
conditions from the same image. \textit{Acc.\ Gap} is the drop from
standard Acc to $P$--$M$--$L$; a model with no relational information
beyond its marginal response bias is expected to drop by 37.5 (chance row). Human results are measured on the 400-instance annotated subset. 
}
\label{tab:relation_accuracy}
\end{table*}

\subsection{Pragmatic Incongruity Detection}
\label{sec:relation_eval}
Table~\ref{tab:relation_accuracy} reports standard accuracy and F1 on the
balanced $P$ versus $M$ comparison. All four models exceed chance, with
LLaVA-v1.6-7B reaching 74.2\% accuracy and 70.8 F1. This range is
consistent with prior zero- and one-shot evaluations of LVLMs on MMSD2.0,
where accuracy has been reported between 46\% and 73\%
\citep{basnet2025evaluating}.

Per-condition accuracy reveals a different pattern: no model exceeds chance on all three conditions ($P$, $M$ and $L$) under zero-shot prompting. InternVL2.5-8B, Qwen2.5-VL-7B
and Idefics2-8B score highly on $M$ (96 to 99\%) but poorly on $P$ (8 to
14\%), consistently defaulting to a negative answer, as reflected in their
low F1 scores (14.0 to 24.3). LLaVA-v1.6-7B shows the opposite tendency,
scoring 62.5\% on $P$ but only 43.1\% on $L$, treating cross-modal
inconsistency as sarcasm regardless of communicative intent. Because standard accuracy evaluates conditions independently, these opposing tendencies remain indistinguishable from genuine relational understanding.

Paired accuracy falls below the 25\%
reference level for three of four models under zero-shot prompting (7.4\%, 13.4\%, and 10.2\%), despite these same models exceeding 96\% accuracy on
$M$ in isolation, indicating that rejecting mismatch carries no information
about whether a model also recognizes incongruity in the same image.
Grouped accuracy never exceeds 26.4\% across all eight configurations, with
three at or below the 12.5\% reference level. LLaVA-v1.6-7B, the strongest
model under standard evaluation, shows the largest gap, falling from 74.2\%
to 19.0\%, a drop that exceeds the 37.5 points expected of a
model with no relational information at all. Standard evaluation thus diverges substantially from relation-level performance, overestimating pragmatic understanding.

\subsubsection{Score-Level Separation Between Relations}
Binary accuracy conflates two failure modes: missing relational
information, or a poor decision threshold applied to good information.
NaturalBench~\citep{li2024naturalbench} shows that adjusting the threshold
alone can raise accuracy by 35 to 40 points, so we compute AUROC over the
continuous sarcasm score (Eq.~1) to separate the two, since it does not
depend on where the threshold falls.

Table~\ref{tab:relation_auroc} shows both cases occur. Qwen2.5-VL-7B
separates $P$ from $M$ well under zero-shot prompting (AUROC 90.0) despite
only 13.4\% paired accuracy, so the score carries information its decision
ignores. InternVL2.5-8B's $P$--$M$ AUROC (66.9) sits closer to chance,
pointing to a weaker representation rather than a threshold problem. $P$--$L$ separation exceeds $P$--$M$ separation for several models (e.g.,
InternVL2.5-8B CoT: 83.7 vs. 63.1), showing that detecting disagreement is
easier than identifying its pragmatic cause. Idefics2-8B's $P$--$L$ AUROC
(14.2) falls below chance, ranking literal pairs as more sarcastic than
sarcastic ones.

CoT prompting lowers $P$--$M$ AUROC for all four models. Since AUROC is threshold-independent, this drop shows that CoT changes the score distribution rather than only the decision boundary, so its accuracy gains in Section~\ref{sec:relation_eval} come partly at
the cost of weaker relational separation~\citep{opitz2024schroedinger}.

\begin{table}[h]
\centering
\small
\resizebox{\linewidth}{!}{
\begin{tabular}{llcc}
\hline
\textbf{Model} & \textbf{Prompt} & \textbf{$P$--$M$} & \textbf{$P$--$L$} \\
\hline
\textit{Random Chance} & & \textit{50.0} & \textit{50.0} \\
\hline
\multirow{2}{*}{InternVL2.5-8B} & Zero-Shot & 66.9 & 82.6 \\
 & CoT & 63.1 & 83.7 \\
\hline
\multirow{2}{*}{Qwen2.5-VL-7B} & Zero-Shot & 90.0 & 62.3 \\
 & CoT & 73.0 & 66.7 \\
\hline
\multirow{2}{*}{LLaVA-v1.6-7B} & Zero-Shot & 85.5 & 51.7 \\
 & CoT & 74.8 & 79.1 \\
\hline
\multirow{2}{*}{Idefics2-8B} & Zero-Shot & 82.5 & 14.2 \\
 & CoT & 61.8 & 53.7 \\
\hline
\end{tabular}}
\caption{
AUROC (\%) separating pragmatic incongruity ($P$) from non-pragmatic
mismatch ($M$) and literal congruity ($L$), computed from the continuous
score (Eq~\ref{eq:score}). 
}
\label{tab:relation_auroc}
\end{table}
\label{sec:auroc}
\subsection{Shortcut Diagnostics}

\begin{table*}[h]
\centering
\small
\resizebox{\linewidth}{!}{
\begin{tabular}{ll|cc|cccc|c|cccc|c|c}
\hline
& & \multicolumn{2}{c|}{\textbf{Baseline}}
& \multicolumn{5}{c|}{\textbf{$P \rightarrow P^-$(masking)}}
& \multicolumn{5}{c|}{\textbf{$M \rightarrow M^+$(injection)}}
& \\
\textbf{Model} & \textbf{Prompt} & \textbf{$P$} & \textbf{$M$}
& \textbf{Lex.} & \textbf{Style} & \textbf{OCR} & \textbf{Avg} & \textbf{Drop}
& \textbf{Lex.} & \textbf{Style} & \textbf{OCR} & \textbf{Avg} & \textbf{Drop}
& $\mathbf{P^-}$--$\mathbf{M^+}$ \\
\hline
\textit{Random Chance} & & \textit{50.0} & \textit{50.0}
& \textit{50.0} & \textit{50.0} & \textit{50.0} & \textit{50.0} & \textit{0.0}
& \textit{50.0} & \textit{50.0} & \textit{50.0} & \textit{50.0} & \textit{0.0}
& \textit{25.0} \\
\hline
\multirow{2}{*}{InternVL2.5-8B} & Zero-Shot
& \phantom{0}9.1 & 98.0
& \phantom{0}5.3 & \phantom{0}6.7 & \phantom{0}4.3 & \phantom{0}5.4 & \textcolor{red}{3.7}
& 91.0 & 94.7 & 89.8 & 91.8 & \textcolor{red}{6.2}
& \phantom{0}5.4 \\
& CoT
& 67.9 & 39.9
& 52.6 & 47.4 & 50.4 & 50.1 & \textcolor{red}{17.8}
& 17.4 & 37.6 & 23.4 & 26.1 & \textcolor{red}{13.8}
& 18.8 \\
\hline
\multirow{2}{*}{Qwen2.5-VL-7B} & Zero-Shot
& 19.3 & 99.3
& 21.1 & 11.1 & \phantom{0}8.7 & 13.6 & \textcolor{red}{5.7}
& 78.9 & 88.3 & 76.1 & 81.1 & \textcolor{red}{18.2}
& \phantom{0}7.7 \\
& CoT
& 77.5 & 61.5
& 63.2 & 59.2 & 62.6 & 61.7 & \textcolor{red}{15.8}
& 79.1 & 72.9 & 80.8 & 77.6 & $-$16.1
& 45.1 \\
\hline
\multirow{2}{*}{LLaVA-v1.6-7B} & Zero-Shot
& 90.2 & 85.9
& 94.7 & 85.9 & 84.0 & 88.2 & \textcolor{red}{2.0}
& 40.4 & 50.7 & \phantom{0}9.4 & 33.5 & \textcolor{red}{52.4}
& 38.2 \\
& CoT
& 95.6 & 29.9
& 94.7 & 93.7 & 91.9 & 93.4 & \textcolor{red}{2.2}
& \phantom{0}1.0 & \phantom{0}3.2 & \phantom{0}0.1 & \phantom{0}1.4 & \textcolor{red}{28.5}
& \phantom{0}4.6 \\
\hline
\multirow{2}{*}{Idefics2-8B} & Zero-Shot
& 14.6 & 96.7
& \phantom{0}5.3 & \phantom{0}6.9 & \phantom{0}6.5 & \phantom{0}6.2 & \textcolor{red}{8.4}
& 96.2 & 94.8 & 96.3 & 95.8 & \textcolor{red}{0.9}
& 15.2 \\
& CoT
& 58.5 & 58.7
& 56.9 & 56.8 & 56.4 & 56.7 & \textcolor{red}{1.8}
& 53.7 & 42.3 & 49.5 & 48.5 & \textcolor{red}{10.2}
& 20.6 \\
\hline
\end{tabular}}
\caption{
Shortcut robustness under masking and injection, by cue type. Baseline
$P$ is measured on the affected pragmatic samples, while baseline $M$ corresponds to Acc $M$ in Table~\ref{tab:relation_accuracy}. \textit{Avg} is the mean across cue types. \textit{Drop} = Baseline $-$ Avg, with positive values (red) indicating accuracy lost under manipulation. Negative values (plain) indicate cases where manipulated accuracy exceeds baseline, suggesting baseline performance was driven by response bias rather than input evidence. $P^-$--$M^+$ is the mean paired accuracy under both
manipulations at once (chance = 25.0).
}
\label{tab:shortcut_robustness}
\end{table*}
\label{sec:shortcut}

Unlike the cross-condition comparisons in Section~\ref{sec:relation_eval}, which may be influenced by stylistic differences between caption sources (Section~\ref{sec:quality_control}), the interventions here edit a single caption in place. With the image, condition, and gold label fixed, any prediction change is attributable to the injected or masked cue.

As established in Section~\ref{sec:benchmark_construction}, lexical, stylistic, and OCR cues are candidate shortcuts: features correlated with labels but not causally related to them. A relation-aware model should remain stable under these perturbations; prediction changes indicate shortcut reliance. Table~\ref{tab:shortcut_robustness} compares $P^-$, $M^+$, and $P^-$--$M^+$ against the affected baseline Acc $P$ and Acc $M$ from Table~\ref{tab:relation_accuracy}.

Injection results show that models are not evaluating the image–caption relation: an unsupported cue should not make an unrelated mismatch sarcastic. Yet LLaVA-v1.6-7B's accuracy on $M$
falls from 85.9 to as low as 9.4 after OCR injection: a few injected words can overturn predictions despite unchanged relations, revealing reliance on cues over relational understanding. InternVL2.5-8B and Idefics2-8B appear more stable under injection
(89.8 to 94.7 and 94.8 to 96.3 against baselines of 98.0 and 96.7), but
this stability does not indicate grounding in the relation either. Both models predict "not sarcastic" for nearly every input (Section~\ref{sec:relation_eval}), leaving little room for sarcasm-inducing cues to change predictions that are already input-independent.
The masking results confirm this: removing cues from sarcastic condition $P$ reduces accuracy to 4.3--6.9, showing that correct predictions relied on cues rather than the image--caption relation. Injection and masking results reveal the same shortcut behavior: models follow cues instead of relations.

Qwen2.5-VL-7B is the only model whose behavior is not fully explained by
this pattern. Under zero-shot prompting its accuracy on $P$ changes little
after masking (8.7 to 21.1 against a 19.3 baseline), while its accuracy on
$M$ still drops substantially after injection (76.1 to 88.3 against a 99.3
baseline), indicating the cue still carries real weight in its decision
even though the model is not wholly dependent on it.

The joint metric $P^-$--$M^+$ is the strictest test, since it requires a
model to reach the correct answer on both conditions at once while the cue evidence has been deliberately set against the true relation. Six of eight configurations score below their own clean $P$--$M$ paired accuracy from Table~\ref{tab:relation_accuracy} under this test, and most fall well below 30, showing that relational discrimination is not robust to manipulable surface cues, indicating that models are not detecting the relation itself. The two exceptions, Qwen2.5-VL-7B under CoT (45.1 vs. 38.3) and Idefics2-8B under zero-shot prompting (15.2 vs. 10.2), arise from strong response bias rather than robustness.

\subsection{Visual Grounding}
\label{sec:modality}

\begin{table}[h]
\centering
\small
\setlength{\tabcolsep}{5pt}
\renewcommand{\arraystretch}{1.05}
\begin{tabular}{ll|c|c}
\hline
\textbf{Model} & \textbf{Input} & \textbf{$P$--$M$} & \textbf{$P^-$--$M^+$} \\
\hline
\textit{Random Chance} & & \textit{25.0} & \textit{25.0} \\
\hline
\multirow{2}{*}{InternVL2.5-8B} & Image+Text & \phantom{0}7.4 & \phantom{0}6.3 \\
 & Text-Only & 47.1 & 43.4 \\
\hline
\multirow{2}{*}{Qwen2.5-VL-7B} & Image+Text & 13.4 & \phantom{0}8.3 \\
 & Text-Only & 55.3 & 28.5 \\
\hline
\multirow{2}{*}{LLaVA-v1.6-7B} & Image+Text & 50.6 & 34.7 \\
 & Text-Only & 71.6 & 53.1 \\
\hline
\multirow{2}{*}{Idefics2-8B} & Image+Text & 10.2 & \phantom{0}9.8 \\
 & Text-Only & \phantom{0}2.6 & 0.0 \\
\hline
\end{tabular}
\caption{\footnotesize
Paired accuracy (zero-shot prompt) with and without the image, under clean
($P$--$M$) and adversarial cue ($P^-$--$M^+$) conditions. The
$P^-$--$M^+$ column averages the lexical and style cues only, since OCR is
undefined without the image.
}
\label{tab:modality_ablation}
\end{table}
Table~\ref{tab:modality_ablation} shows the effect of image removal on relation-level performance. Except for Idefics2-8B, all models perform better with text-only inputs, suggesting that their predictions are driven more by textual cues than by information from the image. The large $P^{-}$--$M^{+}$ drop reveals sensitivity to misleading surface cues. This mirrors caption-only solvability reported for related benchmarks \citep{ouyang2025punchbench, chi2025chimera} and reinforces that condition-level style signal, not visual grounding, drives much of the models' apparent performance. This is the failure mode our perturbation analysis in Section~\ref{sec:shortcut} isolates.
\label{sec:generalisation}

\begin{figure*}[h]
    \centering
    \includegraphics[width=0.6\linewidth]{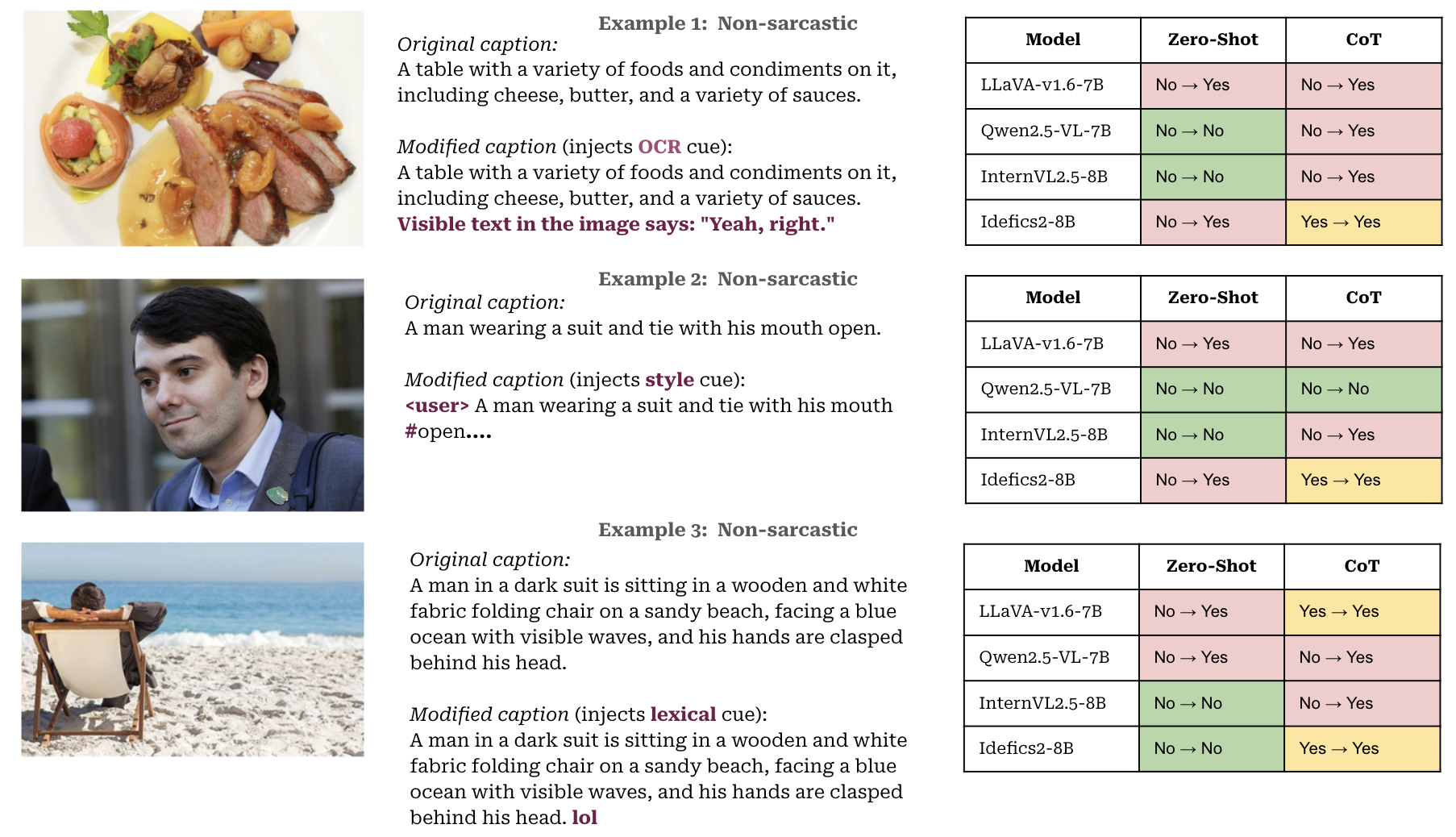}
    \caption{Qualitative examples of shortcut-induced verdict flips across three cue types (OCR, style, lexical) 
    \label{fig:cue2}}
\end{figure*}

\subsection{Model and Data Generalisation}
\label{sec:generalisation_scale_data}
The evaluation above uses four comparable 7--8B models on PragMatch. To assess whether they
are specific to this parameter range or to these model families, we
evaluate Qwen3-VL-32B-Instruct \citep{bai2025qwen3} and Phi-3.5-vision \citep{abdin2024phi3} with the same prompts and metrics. To assess whether they depend on the source split or on the
provenance of the captions, we rebuild the benchmark from the held-out
MMSD2.0 test split (1,000 images, 3,000 pairs) and from SarcNet
\citep{yue-etal-2024-sarcnet} (100 images, 300 pairs) using the same pipeline, and evaluate the same four models on them. Only the source corpus and the caption pool differ, with mismatch captions drawn from the same pool as the literal captions, which are generated with Gemma-3-27B-IT \citep{gemmateam2025gemma3technicalreport}, and hard negatives in SarcNet are less challenging, since they are retrieved from a smaller image pool.  Full results are reported in
Appendix~\ref{app:generalisation}. 

The main findings hold in both checks. Relation-level accuracy stays well
below standard accuracy for the two additional models (86.5\% to 63.7\%
grouped for Qwen3-VL-32B; 74.5\% to 45.5\% for Phi-3.5-vision), and both
remain sensitive to injected cues while losing little accuracy under
masking, with $P^-$--$M^+$ below clean $P$--$M$ in all four of their
configurations. On the two additional sets the response bias of
Section~\ref{sec:relation_eval} recurs: the same three models default to a negative answer and LLaVA-v1.6-7B to a positive one, so this behavior is a property of the model rather than of the source data. Accuracy on $P$ is also higher on MMSD2.0-test than on the main set for every model, which is the opposite of what memorisation of the MMSD2.0 training annotations would predict. The gap between standard and grouped accuracy stays in the same range (23.5 to 51.2 against 30.0 to 55.2 on the main set), the AUROC--accuracy dissociation persists, and injection again overturns predictions. The effects are smaller in absolute terms, and two findings are specific to the main setting: both additional models exceed chance on all three conditions under zero-shot prompting, and the text-only advantage of
Section~\ref{sec:modality} does not transfer. Relation-level evaluation,
shortcut perturbation and the AUROC comparison therefore expose the same
failure modes across model scale and on another data source.

\subsection{Qualitative Examples}

 Figure~\ref{fig:cue2} provides examples of OCR, lexical, and style-based perturbations. Since these perturbations preserve the underlying gold label (Section~\ref{sec:quality_control}), prediction changes indicate shortcut reliance rather than improved multimodal reasoning. We identify distinct failure modes: 1) models treat injected cues as genuine pragmatic evidence: after inserting OCR cue, three CoT models flip, interpreting it as speaker skepticism rather than an artificial cue. 2) style cues cause models to infer unsupported context, here, new speaker intentions despite unchanged semantics. 3) models may correctly recognize incongruity but fail at decision calibration, i.e. model identifies the humorous contrast between formal attire and a beach setting but predicts non-sarcastic until “lol” is added, suggesting that shortcuts act as decision signals rather than evidence. Detailed CoT reasoning outputs are provided in Tables ~\ref{tab:failure1}--\ref{tab:failure3}.
\section{Conclusion}
In this work, we introduce PragMatch, a controlled benchmark for multimodal sarcasm detection that isolates pragmatic incongruity from non-pragmatic mismatch while keeping the image fixed. The benchmark supports relation-level evaluation, controlled shortcut perturbations and modality ablations. Across four comparable 7–8B LVLMs, relation-level accuracy was consistently lower than standard accuracy, showing that strong performance on individual examples did not necessarily translate to consistent predictions across matched image--caption pairs. Shortcut perturbations frequently changed model predictions, and multimodal inputs did not consistently outperform text-only inputs.  These patterns also hold for a larger model, a different model family and two further data sources, so they are not specific to the evaluated models or the source corpus. CoT prompting improves sarcasm recognition for most models but also increases false positive predictions by labeling more non-pragmatic samples as sarcastic. The findings show that current baseline accuracy in multimodal sarcasm understanding can overestimate genuine relational understanding beyond random-chance performance. Therefore, evaluation should be complemented with measures of relational consistency, shortcut robustness and visual grounding.
\section*{Limitations}
A primary limitation of our work is that the benchmark is designed to isolate a specific failure mode, shortcut reliance in multimodal sarcasm detection, rather than provide a comprehensive evaluation of pragmatic understanding.  Although this controlled design enables causal analysis of model behavior, it does not include all shortcut types encountered in natural data and masking is limited to examples containing identifiable cues, as our dataset reveals.  Future work will extend this framework to additional model families, prompting strategies and pragmatic understanding tasks.

Second, our text-only diagnostic reveals that the three conditions differ in caption provenance and style. Sarcastic captions are naturally produced social media text, while literal captions are model-generated visual descriptions. A model could therefore separate $P$ from $L$ partly by concreteness or register alone, a shortcut our benchmark documents but cannot eliminate by design. Conclusions from cross-condition accuracies should be read with this in mind, whereas the perturbation results are unaffected. Constructing literal captions matched to sarcastic captions in abstractness and register is an important extension. More broadly, our three cue families (lexical, stylistic, and OCR) do not exhaust all possible shortcuts, but they target the families most consistently documented in prior sarcasm work, including explicit markers and hashtags \citep{schifanella2016detecting, cai2019multi, qin2023mmsd2} and image-embedded text, which produced the largest effects in our experiments. The benchmark is a starting point covering prevalent, verifiable shortcut types rather than a complete inventory. The OCR cue asserts text absent from the image, testing visual verification rather than invariance. The 25\% and 12.5\% references assume independent uniform guessing and are therefore theoretical rather than empirical.

\section*{Ethics Statement}

All source data in PragMatch is derived from MMSD2.0 \cite{qin2023mmsd2}, an open-source dataset available for academic research, which itself builds on MMSD \cite{cai2019multi}, and from SarcNet \cite{yue-etal-2024-sarcnet} for the generalisation experiments.
We collect no new social media data; user mentions remain anonymized as <user> placeholders following the source dataset. Literal captions are model-generated and describe only observable image content, and injected perturbation cues are limited to benign markers (e.g., "lol", hashtags).
The annotation process was carried out by annotators who are graduate students in university who are English speaking. The instructions they were given are
provided in Appendix~\ref{app:annotation}.
\bibliography{custom}

\appendix
\section*{Appendix}

\section{Prompt Templates}
\label{appendix:prompts}
\subsection{Image Captioning Prompt}
\begin{quote}
\ttfamily
Describe exactly what is visible in the image.\\
Rules:\\
- Only describe visible objects and actions\\
- Do NOT guess context, intention, or events.\\
- If the image contains written text, mention that briefly.\\
- Use one sentence.\\
Now describe the image:
\end{quote}

\subsection{Zero-shot Prompt}

\begin{quote}
\ttfamily
Is the caption sarcastic with respect to the image? Answer only Yes or No.
\end{quote}

\subsection{Zero-shot Text-only Prompt}

\begin{quote}
\ttfamily
Is the text sarcastic? Answer only Yes or No.
\end{quote}

\subsection{Chain-of-Thought Prompt}

\begin{quote}
\ttfamily
You are an expert in multimodal sarcasm detection.\\
Analyze the image and text carefully.\\
1. Describe the image and its meaning in one sentence.\\
2. Describe the text and its meaning in one sentence.\\
3. Explain the overall meaning conveyed when the image and text are interpreted together in one sentence.\\
4. State the author's intended message in one sentence.\\
5. Is the caption sarcastic with respect to the image? Answer only Yes or No.
\end{quote}

\section{Instruction to Annotators}
\label{app:annotation}

\paragraph{Task 1: Sarcasm annotation.}
\begin{enumerate}
\item For each image--caption pair, judge whether the caption is sarcastic
      with respect to the image.
\end{enumerate}

\paragraph{Task 2: Validity check.}
\begin{enumerate}
\setcounter{enumi}{1}
\item For each L caption, judge whether it correctly describes what is
      visible in the image.
\item For each M caption, judge whether it could be read as sarcastic with
      respect to the image.
\end{enumerate}

\paragraph{Task 3: Perturbation validation.}
\begin{enumerate}
\setcounter{enumi}{3}
\item For each edited caption, judge whether the edit changes the gold label.
\item For each edited caption, judge whether the edit changes the pragmatic
      interpretation of the pair.
\end{enumerate}

\paragraph{Potential Risks and Discomfort:} Contents you will see are posts
collected from social media and may include:
\begin{enumerate}
\item Informal or offensive language.
\item References to sensitive topics (e.g., politics, religion, gender).
\item Implicit or explicit stereotypes.
\end{enumerate}

\paragraph{Notes:} If an example is uncomfortable for you to work with, you are allowed to leave it unannotated and end the task at any point.

\paragraph{Compensation:} The annotators are student research assistants employed in the group and paid according to the standard university
pay scale for research assistants.

\section{Generalisation}
\label{app:generalisation}
Tables~\ref{tab:relation_accuracy_app}--\ref{tab:shortcut_robustness_app}
report the configurations summarised in
Section~\ref{sec:generalisation_scale_data}, including the main set evaluated with
Qwen3-VL-32B-Instruct and Phi-3.5-vision-instruct, and the two additional
evaluation sets, namely MMSD2.0-test and SarcNet. Table~\ref{tab:masking_coverage} reports the coverage of each cue type affected by masking across the evaluated datasets.
\begin{table*}[t]
\centering
\begin{tabular}{l|c|c|c}
\hline
Cue type & Main benchmark & MMSD2.0-test & SarcNet \\
\hline
Lexical & 19 (1.9\%)   & 41 (4.1\%)   & 15 (15.0\%) \\
Style   & 638 (63.8\%) & 628 (62.8\%) & 100 (100.0\%) \\
OCR     & 506 (50.6\%) & 765 (76.5\%) & 60 (60.0\%) \\
\hline
\end{tabular}
\caption{Coverage of shortcut cue types among sarcastic examples affected by the masking intervention.}
\label{tab:masking_coverage}
\end{table*}
\begin{table*}[t]
\centering
\resizebox{\linewidth}{!}{
\begin{tabular}{ll|cc|ccc|cc|c}
\hline
& & \multicolumn{2}{c|}{\textbf{Standard Eval.}} & \multicolumn{3}{c|}{\textbf{Per-Condition Acc.}} & \multicolumn{2}{c|}{\textbf{Relation-Level Acc.}} & \\
\textbf{Model} & \textbf{Prompt} & \textbf{Acc} & \textbf{F1} & \textbf{P} & \textbf{M} & \textbf{L} & \textbf{P--M} & \textbf{P--M--L} &
\shortstack[c]{\textbf{Acc. Gap}\\{\scriptsize(Acc $-$ P--M--L)}} \\
\hline
\textit{Random Chance} & & \textit{50.0} & \textit{50.0} & \textit{50.0} & \textit{50.0} & \textit{50.0}
& \textit{25.0} & \textit{12.5} & \textit{37.5} \\
\hline
\multicolumn{10}{l}{\textit{Main benchmark}  (1,000 images, 3,000 pairs)} \\
\hline
\multirow{2}{*}{Qwen3-VL-32B} & Zero-Shot
& 86.5 & 86.5 & 86.6 & 86.3 & 83.2
& 73.9 & 63.7 & \textcolor{red}{22.8} \\
& CoT
& 52.3 & 65.4 & 90.3 & 14.3 & 73.5
& 11.9 & \phantom{0}9.5 & \textcolor{red}{42.8} \\
\hline
\multirow{2}{*}{Phi-3.5-vision-4.2B} & Zero-Shot
& 74.5 & 75.6 & 78.9 & 70.1 & 73.3
& 52.2 & 45.5 & \textcolor{red}{29.0} \\
& CoT
& 55.2 & 44.4 & 35.8 & 74.6 & 73.1
& 19.4 & 16.4 & \textcolor{red}{38.8} \\
\hline
\multicolumn{10}{l}{\textit{MMSD2.0-test} (1,000 images, 3,000 pairs)} \\
\hline
\multirow{2}{*}{InternVL2.5-8B} & Zero-Shot
& 62.6 & 40.9 & 25.9 & 99.4 & 95.0
& 25.3 & 23.0 & \textcolor{red}{39.6} \\
& CoT
& 61.4 & 64.8 & 71.1 & 51.7 & 56.5
& 29.6 & 22.3 & \textcolor{red}{39.1} \\
\hline
\multirow{2}{*}{Qwen2.5-VL-7B} & Zero-Shot
& 65.2 & 47.7 & 31.7 & 98.8 & 82.9
& 31.1 & 19.9 & \textcolor{red}{45.3} \\
& CoT
& 60.5 & 52.2 & 43.2 & 77.7 & 59.9
& 24.6 & 12.9 & \textcolor{red}{47.6} \\
\hline
\multirow{2}{*}{LLaVA-v1.6-7B} & Zero-Shot
& 75.1 & 78.5 & 90.8 & 59.5 & 59.4
& 51.6 & 45.2 & \textcolor{red}{29.9} \\
& CoT
& 53.5 & 68.2 & 99.8 & \phantom{0}7.2 & 10.1
& \phantom{0}7.2 & \phantom{0}2.3 & \textcolor{red}{51.2} \\
\hline
\multirow{2}{*}{Idefics2-8B} & Zero-Shot
& 56.5 & 27.1 & 16.2 & 96.8 & 90.7
& 15.1 & 12.3 & \textcolor{red}{44.2} \\
& CoT
& 52.9 & 29.3 & 19.5 & 86.4 & 78.7
& 12.1 & \phantom{0}7.6 & \textcolor{red}{45.3} \\
\hline
\multirow{2}{*}{Qwen3-VL-32B} & Zero-Shot
& 87.6 & 88.2 & 92.1 & 83.2 & 63.9
& 75.5 & 50.9 & \textcolor{red}{36.7} \\
& CoT
& 60.2 & 69.2 & 89.3 & 31.2 & 51.8
& 24.8 & 14.4 & \textcolor{red}{45.8} \\
\hline
\multirow{2}{*}{Phi-3.5-vision-4.2B} & Zero-Shot
& 67.2 & 71.8 & 83.6 & 50.8 & 53.6
& 37.7 & 31.0 & \textcolor{red}{36.2} \\
& CoT
& 55.0 & 51.8 & 48.3 & 61.7 & 66.7
& 16.7 & 16.7 & \textcolor{red}{38.3} \\
\hline
\multicolumn{10}{l}{\textit{SarcNet} (100 images, 300 pairs)} \\
\hline
\multirow{2}{*}{InternVL2.5-8B} & Zero-Shot
& 59.5 & 31.9 & 19.0 & 100.0 & 100.0
& 19.0 & 19.0 & \textcolor{red}{40.5} \\
& CoT
& 68.2 & 69.3 & 71.9 & 64.6 & 76.0
& 42.7 & 33.3 & \textcolor{red}{34.9} \\
\hline
\multirow{2}{*}{Qwen2.5-VL-7B} & Zero-Shot
& 68.5 & 54.0 & 37.0 & 100.0 & 94.0
& 37.0 & 31.0 & \textcolor{red}{37.5} \\
& CoT
& 62.0 & 49.3 & 37.0 & 87.0 & 75.0
& 27.0 & 18.0 & \textcolor{red}{44.0} \\
\hline
\multirow{2}{*}{LLaVA-v1.6-7B} & Zero-Shot
& 81.5 & 81.0 & 79.0 & 84.0 & 82.0
& 65.0 & 58.0 & \textcolor{red}{23.5} \\
& CoT
& 52.9 & 67.3 & 97.1 & \phantom{0}8.8 & 14.7
& \phantom{0}7.4 & \phantom{0}2.9 & \textcolor{red}{50.0} \\
\hline
\multirow{2}{*}{Idefics2-8B} & Zero-Shot
& 58.0 & 28.8 & 17.0 & 99.0 & 98.0
& 17.0 & 17.0 & \textcolor{red}{41.0} \\
& CoT
& 55.6 & 27.3 & 16.7 & 94.4 & 94.4
& 11.1 & 11.1 & \textcolor{red}{44.4} \\
\hline
\end{tabular}}
\caption{
Standard vs.\ relation-level evaluation for the configurations not shown
in Table~\ref{tab:relation_accuracy}: two additional models on PragMatch,
and all models on the two additional data sources. Columns are as in
Table~\ref{tab:relation_accuracy}. Qwen3-VL-32B-Instruct and
Phi-3.5-vision-instruct were not evaluated on SarcNet.
}
\label{tab:relation_accuracy_app}
\end{table*}

\begin{table}[t]
\centering
\small
\resizebox{\linewidth}{!}{
\begin{tabular}{ll|c|c}
\hline
\textbf{Model} & \textbf{Prompt} & \textbf{$P$--$M$} & \textbf{$P$--$L$} \\
\hline
\textit{Random Chance} & & \textit{50.0} & \textit{50.0} \\
\hline
\multicolumn{4}{l}{\textit{Main benchmark}} \\
\hline
\multirow{2}{*}{Qwen3-VL-32B} & Zero-Shot & 93.5 & 90.5 \\
 & CoT & 73.0 & 89.0 \\
\hline
\multirow{2}{*}{Phi-3.5-vision-4.2B} & Zero-Shot & 81.6 & 80.8 \\
 & CoT & 65.3 & 67.9 \\
\hline
\multicolumn{4}{l}{\textit{MMSD2.0-test}} \\
\hline
\multirow{2}{*}{InternVL2.5-8B} & Zero-Shot & 88.8 & 84.3 \\
 & CoT & 68.9 & 66.7 \\
\hline
\multirow{2}{*}{Qwen2.5-VL-7B} & Zero-Shot & 94.0 & 76.6 \\
 & CoT & 66.6 & 56.0 \\
\hline
\multirow{2}{*}{LLaVA-v1.6-7B} & Zero-Shot & 82.6 & 76.2 \\
 & CoT & 72.1 & 68.4 \\
\hline
\multirow{2}{*}{Idefics2-8B} & Zero-Shot & 77.3 & 74.0 \\
 & CoT & 61.3 & 51.3 \\
\hline
\multirow{2}{*}{Qwen3-VL-32B} & Zero-Shot & 94.8 & 82.9 \\
 & CoT & 75.8 & 74.9 \\
\hline
\multirow{2}{*}{Phi-3.5-vision-4.2B} & Zero-Shot & 75.8 & 71.8 \\
 & CoT & 64.6 & 67.7 \\
\hline
\multicolumn{4}{l}{\textit{SarcNet}} \\
\hline
\multirow{2}{*}{InternVL2.5-8B} & Zero-Shot & 81.7 & 84.9 \\
 & CoT & 77.1 & 80.1 \\
\hline
\multirow{2}{*}{Qwen2.5-VL-7B} & Zero-Shot & 97.4 & 89.7 \\
 & CoT & 73.4 & 65.6 \\
\hline
\multirow{2}{*}{LLaVA-v1.6-7B} & Zero-Shot & 89.2 & 89.9 \\
 & CoT & 77.4 & 73.1 \\
\hline
\multirow{2}{*}{Idefics2-8B} & Zero-Shot & 82.8 & 84.7 \\
 & CoT & 67.7 & 52.9 \\
\hline
\end{tabular}}
\caption{
AUROC (\%) for the configurations not shown in
Table~\ref{tab:relation_auroc}, computed as in that table.
}
\label{tab:relation_auroc_app}
\end{table}

\begin{table}[t]
\centering
\small
\setlength{\tabcolsep}{5pt}
\renewcommand{\arraystretch}{1.05}
\begin{tabular}{ll|c|c}
\hline
\textbf{Model} & \textbf{Input} & \textbf{$P$--$M$} & \textbf{$P^-$--$M^+$} \\
\hline
\textit{Random Chance} & & \textit{25.0} & \textit{25.0} \\
\hline
\multicolumn{4}{l}{\textit{Main benchmark}} \\
\hline
\multirow{2}{*}{Qwen3-VL-32B} & Image+Text & 73.9 & 46.0 \\
 & Text-Only & 53.8 & 49.7 \\
\hline
\multirow{2}{*}{Phi-3.5-vision-4.2B} & Image+Text & 52.2 & 40.5 \\
 & Text-Only & 31.4 & 37.1 \\
\hline
\multicolumn{4}{l}{\textit{MMSD2.0-test}} \\
\hline
\multirow{2}{*}{InternVL2.5-8B} & Image+Text & 25.3 & 24.6 \\
 & Text-Only & 17.3 & 21.1 \\
\hline
\multirow{2}{*}{Qwen2.5-VL-7B} & Image+Text & 31.1 & 34.7 \\
 & Text-Only & \phantom{0}9.5 & \phantom{0}8.4 \\
\hline
\multirow{2}{*}{LLaVA-v1.6-7B} & Image+Text & 51.6 & 40.8 \\
 & Text-Only & 49.4 & 33.6 \\
\hline
\multirow{2}{*}{Idefics2-8B} & Image+Text & 15.1 & 16.7 \\
 & Text-Only & 25.0 & 16.6 \\
\hline
\multirow{2}{*}{Qwen3-VL-32B} & Image+Text & 75.5 & 64.7 \\
 & Text-Only & 44.2 & 41.7 \\
\hline
\multirow{2}{*}{Phi-3.5-vision-4.2B} & Image+Text & 37.7 & 31.6 \\
 & Text-Only & 31.6 & 34.8 \\
\hline
\multicolumn{4}{l}{\textit{SarcNet}} \\
\hline
\multirow{2}{*}{InternVL2.5-8B} & Image+Text & 19.0 & 22.5 \\
 & Text-Only & 26.0 & 37.3 \\
\hline
\multirow{2}{*}{Qwen2.5-VL-7B} & Image+Text & 37.0 & 49.3 \\
 & Text-Only & 24.0 & 42.0 \\
\hline
\multirow{2}{*}{LLaVA-v1.6-7B} & Image+Text & 65.0 & 69.3 \\
 & Text-Only & 63.0 & 54.3 \\
\hline
\multirow{2}{*}{Idefics2-8B} & Image+Text & 17.0 & 12.7 \\
 & Text-Only & 38.0 & 37.2 \\
\hline
\end{tabular}
\caption{\footnotesize
Paired accuracy (zero-shot prompt) for the configurations not shown in
Table~\ref{tab:modality_ablation}, under clean ($P$--$M$) and adversarial
cue ($P^-$--$M^+$) conditions, excluding OCR.
}
\label{tab:modality_ablation_app}
\end{table}

\begin{table*}[t]
\centering
\small
\setlength{\tabcolsep}{3pt}
\renewcommand{\arraystretch}{0.9}

(a) \textit{Main benchmark}\\[2pt]
\begin{tabular}{ll|c|c|c}
\hline
Model & Prompt & P--M & P--M--L & AUROC P--M \\
\hline
InternVL2.5-8B & Zero-shot & 7.4 [5.9, 9.1] & 6.0 [4.6, 7.6] & 66.9 [64.3, 69.1] \\
& CoT & 25.8 [23.1, 28.5] & 23.3 [20.6, 25.8] & 63.1 [60.9, 65.2] \\
 
\hline
Qwen2.5-VL-7B & Zero-shot & 13.4 [11.3, 15.6] & 5.6 [4.3, 7.0] & 90.0 [88.8, 91.2] \\
& CoT & 38.3 [35.2, 41.3] & 26.4 [23.8, 29.1] & 73.0 [71.3, 74.7] \\
 
\hline
LLaVA-1.6-7B & Zero-shot & 50.6 [47.6, 53.4] & 19.0 [16.5, 21.3] & 85.5 [83.9, 87.0] \\
& CoT & 28.1 [25.2, 30.9] & 17.4 [15.1, 19.8] & 74.8 [72.8, 76.8] \\
 
\hline
Idefics2-8B & Zero-shot & 10.2 [8.3, 12.3] & 2.7 [1.8, 3.8] & 82.5 [80.8, 84.3] \\
& CoT & 28.8 [25.7, 31.5] & 15.8 [13.7, 17.9] & 61.8 [59.7, 63.8] \\
 
\hline
Qwen3-VL-32B & Zero-shot & 73.9 [71.4, 76.4] & 63.7 [60.8, 66.8] & 93.5 [92.6, 94.5] \\
& CoT & 11.9 [9.9, 13.9] & 9.5 [7.8, 11.2] & 73.0 [70.8, 75.0] \\

\hline
Phi-3.5-vision-4.2B & Zero-shot & 52.2 [49.1, 55.3] & 45.5 [42.4, 48.5] & 81.6 [79.9, 83.3] \\
& CoT & 19.4 [14.8, 24.6] & 16.4 [9.0, 25.4] & 65.3 [58.7, 72.5] \\
 
\bottomrule
\end{tabular}

\vspace{6pt}

(b) \textit{MMSD2.0-test}\\[2pt]
\begin{tabular}{ll|c|c|c}
\hline
Model & Prompt & P--M & P--M--L & AUROC P--M \\
\hline
InternVL2.5-8B & Zero-shot & 25.3 [22.6, 28.0] & 23.0 [20.4, 25.5] & 88.8 [87.5, 90.1] \\
& CoT & 29.6 [26.8, 32.5] & 22.3 [19.8, 24.9] & 68.9 [67.1, 70.8] \\
 
\hline
Qwen2.5-VL-7B & Zero-shot & 31.1 [28.4, 34.1] & 19.9 [17.4, 22.5] & 94.0 [93.0, 95.0] \\
& CoT & 24.6 [21.9, 27.4] & 12.9 [10.9, 14.9] & 66.6 [65.0, 68.3] \\
 
\hline
LLaVA-1.6-7B & Zero-shot & 51.6 [48.7, 54.4] & 45.2 [42.1, 48.1] & 82.6 [81.0, 84.1] \\
& CoT & 7.2 [5.5, 8.8] & 2.3 [1.5, 3.3] & 72.1 [69.7, 74.5] \\
 
\hline
Idefics2-8B & Zero-shot & 15.1 [12.8, 17.3] & 12.3 [10.1, 14.5] & 77.3 [75.3, 79.2] \\
& CoT & 12.1 [10.0, 14.4] & 7.6 [5.9, 9.5] & 61.3 [59.8, 63.2] \\
 
\hline
Qwen3-VL-32B & Zero-shot & 75.5 [72.7, 78.1] & 50.9 [47.8, 54.0] & 94.8 [93.9, 95.7] \\
& CoT & 24.8 [22.2, 27.6] & 14.4 [12.3, 16.6] & 75.8 [73.9, 77.7] \\

\hline
Phi-3.5-vision-4.2B & Zero-shot & 37.7 [34.6, 40.7]  & 31.0 [28.1, 33.9] & 75.8 [74.0, 77.5] \\
& CoT & 16.7 [8.3, 26.7] & 16.7 [8.3, 26.7] & 64.6 [56.8, 71.9] \\
 
\hline
\end{tabular}

\vspace{6pt}

(c) \textit{SarcNet}\\[2pt]
\begin{tabular}{ll|c|c|c}
\hline
Model & Prompt & P--M & P--M--L & AUROC P--M \\
\hline
InternVL2.5-8B & Zero-Shot
& 19.0 [12.0, 28.0] & 19.0 [12.0, 28.0] & 81.7 [76.3, 87.0] \\
& CoT
& 42.7 [33.3, 52.1] & 33.3 [24.0, 42.7] & 77.1 [71.6, 82.7] \\
\hline
Qwen2.5-VL-7B & Zero-Shot
& 37.0 [28.0, 46.0] & 31.0 [22.0, 39.0] & 97.4 [95.5, 98.9] \\
& CoT
& 27.0 [19.0, 36.0] & 18.0 [11.0, 26.0] & 73.4 [68.3, 78.8] \\
\hline
LLaVA-v1.6-7B & Zero-Shot
& 65.0 [56.0, 75.0] & 58.0 [48.0, 68.0] & 89.2 [84.7, 93.4] \\
& CoT
& 7.4 [1.5, 14.7] & 2.9 [0.0, 7.4] & 77.4 [70.5, 83.8] \\
\hline
Idefics2-8B & Zero-Shot
& 17.0 [10.0, 24.0] & 17.0 [10.0, 24.0] & 82.8 [77.5, 87.4] \\
& CoT
& 11.1 [2.8, 22.2] & 11.1 [2.8, 22.2] & 67.7 [60.2, 76.3] \\
\hline
\end{tabular}

\caption[Relation-level accuracy (\%) and AUROC (\%) with 95\% CI]{Relation-level accuracy (\%) and AUROC (\%) with 95\% confidence intervals. Intervals are block bootstrap intervals over images (1,000 resamples), keeping all conditions of an image together.}
\label{tab:pragmatch_mmsd_confidence}
\end{table*}

\begin{table*}[h]
\centering
\small
\resizebox{\linewidth}{!}{
\begin{tabular}{ll|cc|cccc|c|cccc|c|c}
\hline
& & \multicolumn{2}{c|}{\textbf{Baseline}}
& \multicolumn{5}{c|}{\textbf{$P \rightarrow P^-$(masking)}}
& \multicolumn{5}{c|}{\textbf{$M \rightarrow M^+$(injection)}}
& \\
\textbf{Model} & \textbf{Prompt} & \textbf{$P$} & \textbf{$M$}
& \textbf{Lex.} & \textbf{Style} & \textbf{OCR} & \textbf{Avg} & \textbf{Drop}
& \textbf{Lex.} & \textbf{Style} & \textbf{OCR} & \textbf{Avg} & \textbf{Drop}
& $\mathbf{P^-}$--$\mathbf{M^+}$ \\
\hline
\textit{Random Chance} & & \textit{50.0} & \textit{50.0}
& \textit{50.0} & \textit{50.0} & \textit{50.0} & \textit{50.0} & \textit{0.0}
& \textit{50.0} & \textit{50.0} & \textit{50.0} & \textit{50.0} & \textit{0.0}
& \textit{25.0} \\
\hline
\multicolumn{15}{l}{\textit{Main benchmark}} \\
\hline
\multirow{2}{*}{Qwen3-VL-32B} & Zero-Shot
& 87.4 & 86.3
& 73.7 & 86.4 & 86.0 & 82.0 & \textcolor{red}{5.4}
& 23.9 & 84.2 & 63.5 & 57.2 & \textcolor{red}{29.1}
& 51.3 \\
& CoT
& 85.8 & 14.3
& 78.9 & 85.3 & 86.2 & 83.5 & \textcolor{red}{2.3}
& \phantom{0}6.4 & 15.9 & \phantom{0}7.5 & \phantom{0}9.9 & \textcolor{red}{4.4}
& \phantom{0}5.6 \\
\hline
\multirow{2}{*}{Phi-3.5-vision-4.2B} & Zero-Shot
& 83.1 & 70.1
& 73.7 & 80.6 & 79.4 & 77.9 & \textcolor{red}{5.2}
& 11.9 & 80.9 & \phantom{0}0.5 & 31.1 & \textcolor{red}{39.0}
& 27.1 \\
& CoT
& 59.6 & 74.7
& \phantom{0}0.0 & 50.0 & 38.1 & 29.4 & \textcolor{red}{30.2}
& 60.8 & 77.4 & \phantom{0}7.3 & 48.5 & \textcolor{red}{26.2}
& 10.0 \\
\hline
\multicolumn{15}{l}{\textit{MMSD2.0-test}} \\
\hline
\multirow{2}{*}{InternVL2.5-8B} & Zero-Shot
& 23.1 & 99.4
& 29.3 & 28.2 & 22.9 & 26.8 & $-$3.7
& 98.9 & 99.0 & 97.5 & 98.5 & \textcolor{red}{0.9}
& 23.8 \\
& CoT
& 75.4 & 51.7
& 75.6 & 71.9 & 64.4 & 70.6 & \textcolor{red}{4.8}
& 32.0 & 44.3 & 36.7 & 37.7 & \textcolor{red}{14.0}
& 20.6 \\
\hline
\multirow{2}{*}{Qwen2.5-VL-7B} & Zero-Shot
& 31.9 & 98.8
& 36.6 & 33.4 & 23.4 & 31.1 & \textcolor{red}{0.8}
& 99.2 & 98.7 & 90.9 & 96.3 & \textcolor{red}{2.5}
& 30.1 \\
& CoT
& 49.5 & 77.7
& 58.5 & 39.8 & 46.8 & 48.4 & \textcolor{red}{1.1}
& 75.3 & 74.1 & 73.6 & 74.3 & \textcolor{red}{3.4}
& 26.5 \\
\hline
\multirow{2}{*}{LLaVA-v1.6-7B} & Zero-Shot
& 94.1 & 59.5
& 95.1 & 91.1 & 90.8 & 92.4 & \textcolor{red}{1.7}
& 38.1 & 60.9 & \phantom{0}1.0 & 33.3 & \textcolor{red}{26.2}
& 27.2 \\
& CoT
& 99.9 & \phantom{0}7.2
& 100.0 & 99.7 & 99.2 & 99.6 & \textcolor{red}{0.3}
& \phantom{0}1.2 & \phantom{0}5.7 & \phantom{0}0.1 & \phantom{0}2.3 & \textcolor{red}{4.9}
& \phantom{0}1.9 \\
\hline
\multirow{2}{*}{Idefics2-8B} & Zero-Shot
& 17.1 & 96.8
& 17.1 & 16.9 & 12.9 & 15.6 & \textcolor{red}{1.4}
& 97.6 & 98.0 & 97.4 & 97.7 & $-$0.9
& 15.4 \\
& CoT
& 21.6 & 86.4
& 33.3 & \phantom{0}7.8 & 18.1 & 19.7 & \textcolor{red}{1.9}
& 86.3 & 88.1 & 79.5 & 84.6 & \textcolor{red}{1.7}
& 18.4 \\
\hline
\multirow{2}{*}{Qwen3-VL-32B} & Zero-Shot
& 93.0 & 83.2
& 97.6 & 92.2 & 85.5 & 91.7 & \textcolor{red}{1.3}
& 41.2 & 88.2 & 59.3 & 62.9 & \textcolor{red}{20.3}
& 61.2 \\
& CoT
& 88.5 & 31.2
& 92.7 & 88.5 & 91.1 & 90.8 & $-$2.3
& 15.6 & 29.9 & 15.9 & 20.5 & \textcolor{red}{10.7}
& 21.2 \\
\hline
\multirow{2}{*}{Phi-3.5-vision-4.2B} & Zero-Shot
& 85.8 & 50.8
& 90.2 & 85.0 & 78.2 & 84.5 & \textcolor{red}{1.3}
& 23.0 & 59.0 & \phantom{0}0.0 & 27.3 & \textcolor{red}{23.5}
& 21.1 \\
& CoT
& 34.1 & 66.9
& \phantom{0}0.0 & 46.9 & 45.5 & 30.8 & \textcolor{red}{3.3}
& 70.6 & 61.1 & 15.6 & 49.1 & \textcolor{red}{17.8}
& 13.2 \\
\hline
\multicolumn{15}{l}{\textit{SarcNet}} \\
\hline
\multirow{2}{*}{InternVL2.5-8B} & Zero-Shot
& 19.0 & 100.0
& 20.0 & 25.0 & 18.3 & 21.1 & $-$2.1
& 100.0 & 100.0 & 97.0 & 99.0 & \textcolor{red}{1.0}
& 20.6 \\
& CoT
& 71.9 & 64.6
& 66.7 & 66.7 & 66.1 & 66.5 & \textcolor{red}{5.4}
& 38.5 & 57.9 & 45.8 & 47.4 & \textcolor{red}{17.2}
& 36.9 \\
\hline
\multirow{2}{*}{Qwen2.5-VL-7B} & Zero-Shot
& 48.4 & 100.0
& 66.7 & 32.0 & 31.7 & 43.4 & \textcolor{red}{5.0}
& 100.0 & 100.0 & 96.0 & 98.7 & \textcolor{red}{1.3}
& 42.9 \\
& CoT
& 39.6 & 87.0
& 40.0 & 35.0 & 36.7 & 37.2 & \textcolor{red}{2.3}
& 91.0 & 89.0 & 88.0 & 89.3 & $-$2.3
& 34.3 \\
\hline
\multirow{2}{*}{LLaVA-v1.6-7B} & Zero-Shot
& 84.1 & 84.0
& 80.0 & 83.0 & 88.3 & 83.8 & \textcolor{red}{0.3}
& 69.0 & 89.0 & \phantom{0}5.0 & 54.3 & \textcolor{red}{29.7}
& 46.8 \\
& CoT
& 97.9 & \phantom{0}8.5
& 100.0 & 95.5 & 100.0 & 98.5 & $-$0.6
& \phantom{0}4.5 & 10.6 & \phantom{0}0.0 & \phantom{0}5.1 & \textcolor{red}{3.5}
& \phantom{0}3.1 \\
\hline
\multirow{2}{*}{Idefics2-8B} & Zero-Shot
& 17.0 & 99.0
& 13.3 & 12.0 & 15.0 & 13.4 & \textcolor{red}{3.6}
& 100.0 & 100.0 & 98.0 & 99.3 & $-$0.3
& 12.9 \\
& CoT
& 16.7 & 94.4
& 25.0 & 15.4 & 27.3 & 22.6 & $-$5.9
& 97.1 & 100.0 & 91.2 & 96.1 & $-$1.7
& 19.5 \\
\hline
\end{tabular}}
\caption{
Shortcut robustness for the configurations not shown in
Table~\ref{tab:shortcut_robustness}. Columns are as in that table.
}
\label{tab:shortcut_robustness_app}
\end{table*}

\begin{table*}[h]
\centering
\small
\setlength{\tabcolsep}{3pt}
\caption{Shortcut injection transformations used to probe reliance on superficial textual cues. Lexical and style cues are inserted to the caption, while OCR cues are appended as simulated text extracted from the image.}
\begin{tabular}{p{0.9cm}| p{5.2cm} |p{6.2cm}}
\hline
\textbf{Cue} & \textbf{Transformation}  & \textbf{Example} \\
\hline

Lexical &
Discourse markers, sarcasm cues, and informal expressions
(\texttt{lol}, \texttt{haha}, \texttt{lmao}, \texttt{ironic}, \texttt{smh},
\texttt{wow}, \texttt{great}) 

&
"The weather is nice." $\rightarrow$
"The weather is nice. lol" \\

\hline

Style &
Formatting and stylistic modifications: \texttt{<user>}, hashtags,
\textsc{caps}, ellipsis (\texttt{...}), and multiple exclamation marks
(\texttt{!!!})
&
"The weather is nice." $\rightarrow$
"<user> The weather is \#nice..." \\

\hline

OCR &
Caption-level: append a claim about visible text
&
"The weather is nice." $\rightarrow$
"The weather is nice. Visible text in the image says:
`Yeah, right.' " \\

\hline
\end{tabular}
\label{tab:injection_temp}
\end{table*}
\begin{table*}[h]
\centering
\scriptsize
\setlength{\tabcolsep}{3pt}
\caption{Example 1: OCR cue. Three models treat the injected phrase "Yeah, right." as genuine first-order speaker evidence rather than as a perturbation disconnected from the underlying image content, each explicitly naming skepticism, irony, or dismissiveness as the basis for the sarcastic verdict.}
\label{tab:failure1}
\begin{tabular}{p{2.2cm} p{1.3cm} p{1.2cm} p{9.5cm}}
\hline
\textbf{Model} & \textbf{Caption} & \textbf{Verdict} & \textbf{Quoted reasoning} \\
\hline
LLaVA-v1.6-7B-CoT &
Modified &
Yes &
"implies skepticism or disbelief towards the image, possibly suggesting that the food depicted is not as healthy or nutritious as it appears"  \\
\hline
Qwen2.5-VL-7B-CoT &
Modified &
Yes &
"suggests skepticism or irony towards the food presentation"  \\
\hline
InternVL2.5-8B-CoT &
Modified &
Yes &
"is a dismissive expression, suggesting skepticism or disbelief"   \\
\hline
\end{tabular}
\end{table*}

\begin{table*}[h]
\centering
\scriptsize
\setlength{\tabcolsep}{3pt}
\caption{Example 2: Style cue. LLaVA-CoT's baseline reasoning assigns the caption no specific meaning, then constructs a new "person commenting" framing once only formatting markers are added. InternVL-CoT's own baseline reasoning already names the same mouth-open/image discrepancy that its modified pass later cites as new evidence, confirming the discrepancy was present before the intervention. }
\label{tab:failure2}
\begin{tabular}{p{2.2cm} p{1.3cm} p{1.2cm} p{9.5cm}}
\hline
\textbf{Model} & \textbf{Caption} & \textbf{Verdict} & \textbf{Quoted reasoning} \\
\hline
LLaVA-v1.6-7B-CoT &
Baseline &
No &
"The text is a description of the image, but without additional context, it does not provide any specific meaning."   \\
\hline
LLaVA-v1.6-7B-CoT &
Modified &
Yes &
"A person commenting on a photo of a man wearing a suit and tie with his mouth open."  \\
\hline
InternVL2.5-8B-CoT &
Baseline &
No &
"The intended message is to highlight the contrast between the man's actual expression and the text's description."   \\
\hline
InternVL2.5-8B-CoT &
Modified &
Yes &
"The text describes a man wearing a suit and tie with his mouth open, which is not visible in the image."  \\
\hline
\end{tabular}
\end{table*}

\begin{table*}[h]
\centering
\scriptsize
\setlength{\tabcolsep}{3pt}
\caption{Example 3: Lexical cue. Qwen-CoT's own baseline reasoning already identifies the sarcastic mechanism (the businessman/beach contrast) yet its baseline verdict remains "No"; appending the lexical marker "lol" flips the verdict to "Yes" with no change to the image--text relationship, indicating the shortcut acts as a confidence threshold rather than new evidence. LLaVA-CoT and Idefics-CoT predict "Yes" correctly from the outset, and remain unchanged after the lexical intervention. }
\label{tab:failure3}
\begin{tabular}{p{2.2cm} p{1.3cm} p{1.2cm} p{9.5cm}}
\hline
\textbf{Model} & \textbf{Caption} & \textbf{Verdict} & \textbf{Quoted reasoning} \\
\hline
Qwen2.5-VL-7B-CoT &
Baseline &
No &
"The juxtaposition of a businessman relaxing on a beach creates a humorous contrast between work and leisure."   \\
\hline
Qwen2.5-VL-7B-CoT &
Modified &
Yes &
Verdict flips after "lol" is appended; the image--text relationship is otherwise unchanged.   \\
\hline
LLaVA-v1.6-7B-CoT &
Baseline &
Yes &
"The image and text together create a contrast between the formality of the man's clothing and the casual setting of the [\ldots]"   \\
\hline
LLaVA-v1.6-7B-CoT &
Modified &
Yes &
Verdict unchanged; same grounding carries through the lexical intervention. \\
\hline
Idefics2-8B-CoT &
Baseline &
Yes &
"A man is sitting on a beach chair on the sand facing the ocean."   \\
\hline
Idefics2-8B-CoT &
Modified &
Yes &
Verdict unchanged; same grounding carries through the lexical intervention. \\
\hline
\end{tabular}
\end{table*}

\end{document}